\documentclass{article}

 \usepackage[final,sglblindworkshop]{neurips_2025}
 \workshoptitle{Mathematical Foundations and Operational Integration of ML for Uncertainty-Aware Decision-Making}

\usepackage{graphicx} 
\usepackage{microtype}
\usepackage{graphicx}
\usepackage{subfigure}
\usepackage{booktabs}
 \usepackage{neurips_2025}
 \usepackage{wrapfig}
 \usepackage{subcaption}
\usepackage[utf8]{inputenc} 
\usepackage[T1]{fontenc}    
\usepackage{hyperref}       
\usepackage{url}            
\usepackage{booktabs}       
\usepackage{amsfonts}       
\usepackage{nicefrac}       
\usepackage{microtype}      
\usepackage{xcolor}   
\usepackage{neurips_2025}
\usepackage{graphicx} 
\usepackage{microtype}
\usepackage{graphicx}
\usepackage{subfigure}
\usepackage{booktabs} 

\usepackage[utf8]{inputenc} 
\usepackage[T1]{fontenc}    
\usepackage{hyperref}       
\usepackage{url}            
\usepackage{booktabs}       
\usepackage{amsfonts}       
\usepackage{nicefrac}       
\usepackage{microtype}      
\usepackage{xcolor}         

\usepackage{amsmath}
\usepackage{listings}
\usepackage{float}
\usepackage{changepage}
\usepackage{tabularx}
\usepackage{graphicx}

\title{Can Linear Probes Measure LLM Uncertainty ?}

\author{%
   Ramzi Dakhmouche \thanks{Equal contribution} \\
{  \scriptsize{Institute of Mathematics, EPFL, Switzerland }}\\
 { \scriptsize Computational Engineering Lab, Empa, Switzerland} \\
\scriptsize  \texttt{ramzi.dakhmouche@epfl.ch} \\
  \And
  Adrien Letellier $^{*}$ \\
{ \scriptsize ENSAE, IP Paris, France }\\
 {  \scriptsize Computational Engineering Lab, Empa, Switzerland }\\
\scriptsize \texttt{adrien.letellier@ensae.fr} \\
  \AND
 Hossein Gorji \\
  {\scriptsize{ Computational Engineering Lab, Empa, Switzerland }}
 \\
 \scriptsize \texttt{mohammadhossein.gorji@empa.ch} \\
}

\begin{document}

\maketitle

\begin{abstract}
Effective Uncertainty Quantification (UQ) represents a key aspect for reliable deployment of Large Language Models (LLMs) in automated decision-making and beyond. Yet, for LLM generation with structured output, the state-of-the-art in UQ is still dominated by the naive baseline given by the maximum softmax score. To address this shortcoming, we demonstrate that taking a principled approach via Bayesian statistics leads to improved performance despite leveraging the simplest possible model, namely linear regression. More precisely, we propose to train multiple Bayesian linear models, each predicting the output of a layer given the output of the previous one. Based on the obtained layer-level posterior distributions, we infer the global uncertainty level of the LLM by identifying a sparse combination of distributional features, leading to an efficient UQ scheme. Numerical experiments on various LLMs show consistent improvement over state-of-the-art baselines. 
\end{abstract}

\section{Introduction}


Designing efficient Uncertainty Quantification (UQ) schemes for Large Language Models (LLMs) is a key requirement for their deployment in practical decision-making settings. A very common setup, for the latter, involves choosing the optimal option among a number of possible choices, such as in investment portfolio selection based on unstructured data from news articles, legal filings, or audit reports. LLMs increasingly act as recommender systems filtering and ranking best options, while incorporating heterogeneous information sources.

Yet, in discrete choice settings, existing UQ approaches fail to beat simple baselines \cite{Vashurin_2025}. Indeed, the current best performing UQ method for LLMs consists of using the highest softmax score given by the LLM's last layer. It is worth noting though that, most existing approaches are not based on a principled Bayesian statistics framework but rather on post-processing of soft-max scores \cite{kuhn2023semantic,lin2024generating,duan2024shifting, farquhar2024detecting}. For instance, \cite{farquhar2024detecting, kuhn2023semantic} propose to cluster answers to open questions into classes based on similar meaning, then compute the entropy of the distribution over classes by aggregating softmax scores. 

The lack of principled factual UQ approaches for LLMs has been mostly due to the untractable nature of Bayesian inference for large-scale neural networks. However, recent work on LLM interpretability \cite{belrose2023eliciting, halawioverthinking, dar2023analyzing} suggest that much of the LLM's intermediate processing can be well approximated by trained linear maps. This raises the question of whether one could leverage such approximations to design an efficient UQ estimation approach based on a Bayesian framework. We investigate this question and provide a positive answer by making the following contributions: 
\begin{itemize}
    \item We design, Bayesian Linear Lens (BLL), a UQ approach approximating the posterior distributions of LLM layers, and aggregating them via a sparse combination to obtain global UQ estimates,
    \item We reduce the dimensionality of the statistics defining the posteriors to focus the estimation on the most relevant components, while improving the scalability of the approach, 
    \item We compare against state-of-the-art baselines demonstrating consistent performance improvement. 
\end{itemize}

\section{Background \& Problem Setting}
\subsection{Background}
\begin{wrapfigure}{r}{0.5\textwidth}  
    \includegraphics[scale=0.4]{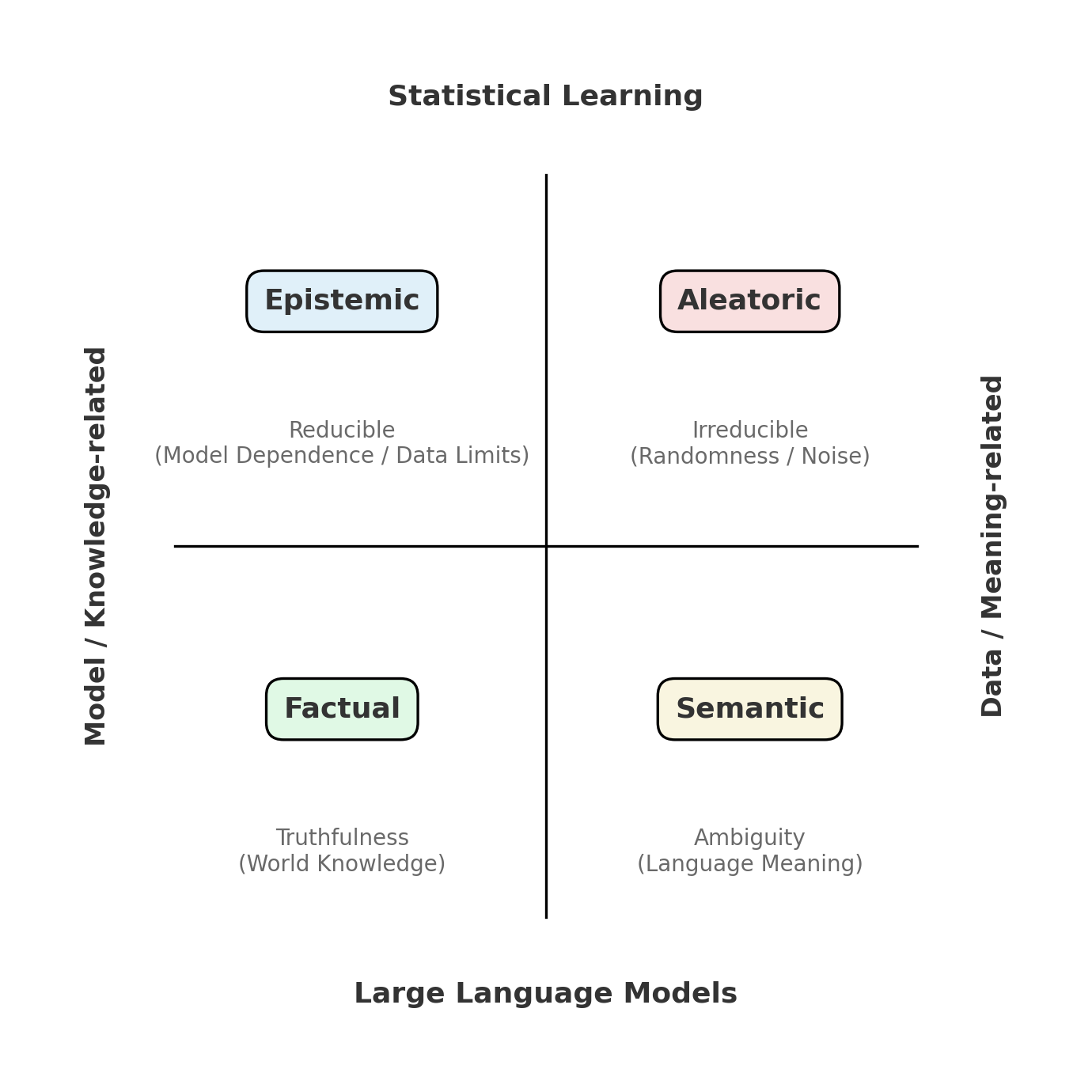}
    \caption{UQ Decomposition in Learning Systems }
    \label{fig:sin}
\end{wrapfigure}
In uncertainty quantification, it is common \cite{kendall2017uncertainties} to distinguish between \emph{epistemic uncertainty}, which arises from limited knowledge or imperfect models and can in principle be reduced with more data or better modeling, and \emph{aleatoric uncertainty}, which captures the inherent randomness of the data-generating process and is irreducible. 
In the context of LLMs, a related dichotomy has emerged between factual uncertainty, which concerns uncertainty about the truthfulness of generated content relative to external world knowledge, and semantic uncertainty, which captures ambiguity or multiplicity in the possible meanings and continuations of a prompt. While the former aligns closely with epistemic uncertainty, the latter reflects linguistic indeterminacy inherent to natural language and is thus more akin to aleatoric variability. We focus in this work on factual UQ given its major role in automated decision making. Hence, we measure the LLM uncertainty in its response to Multiple Choice Questions as a first step, to disentangle semantic and  factual uncertainties. 
\subsection{Bayesian Linear Regression}
We recall the Bayesian linear estimation setting which we leverage in the following sections. 
Consider a standard linear model with inputs $x \in \mathbb{R}^d$ and outputs $y \in \mathbb{R}$:
\[
y = x^\top w + \varepsilon, \quad \varepsilon \sim \mathcal{N}(0, \sigma^2),
\]
where $w \in \mathbb{R}^d$ are the regression weights and $\sigma^2$ is the observation noise variance. In the Bayesian setting, we place a prior distribution on the weights, which we choose to be Gaussian for simplicity and tractability of the posterior:
\[
w \sim \mathcal{N}(0, \tau^2 I_d).
\]

Given a dataset $D = \{(x_i, y_i)\}_{i=1}^n$, the likelihood is given by:
$
p(y \mid X, w) = \mathcal{N}(y \mid Xw, \sigma^2 I_n),
$ where $X \in \mathbb{R}^{n \times d}$ is the design matrix and $y \in \mathbb{R}^n$ the stacked responses. 

By conjugacy, the posterior over $w$ is also Gaussian:
\[
p(w \mid X, y) = \mathcal{N}(w \mid \mu_n, \Sigma_n),
\]
with
\[
\Sigma_n = \left(\tfrac{1}{\sigma^2} X^\top X + \tfrac{1}{\tau^2} I_d \right)^{-1}, 
\quad
\mu_n = \tfrac{1}{\sigma^2} \Sigma_n X^\top y.
\]

For a new test input $x_\ast$, the posterior predictive distribution has the following closed-form:
\[
p(y_\ast \mid x_\ast, X, y) = \mathcal{N}\!\left(x_\ast^\top \mu_n,\; x_\ast^\top \Sigma_n x_\ast + \sigma^2\right).
\]

\section{Bayesian Linear Lens}
\label{sec:bayesian}

Motivated by interpretability results \cite{belrose2023eliciting, lindsey2025biology} showing that various LLM layers are mostly deactivated when the LLM is hallucinating, making the corresponding hidden states linearly predictable, we design a simple yet effective UQ method to capture factual uncertainty via Bayesian linear regression. More precisely, given a LLM with $L$ layers and  hidden size $D$, denote by
$u \in \{0, 1\}$ the truthfulness of the LLM for a given  question, $h^{(l, i)}$ the neuron $i$ at layer $l$ and $y^{(l, i)} = h^{(l,i)} - h^{(l-1, i)}$ the neuron's \emph{centered} activation. We are interested in estimating $p(y^{(l, i)}|u=1)$ and $p(y^{(l, i)}|u=0)$, the posterior distributions, derived from linear models, corresponding to training subsets where the LLM responded correctly and incorrectly, respectively. For that matter, we take the following steps:

\begin{enumerate}
    \item We collect sufficient statistics\footnote{A sufficient statistic is a summary of the data that keeps all the information needed to learn about the unknown quantity, here the posterior of each layer. In the notation of section 2.2, they would correspond to $X^\top X$ and $X^{\top}y$.} on a training dataset. We distinguish between samples corresponding to correct and incorrect answers of the LLM, hereafter dubbed Cor and Incor. Furthermore, we either consider the hidden states of the generated token  or its average with the hidden states of the question: $h^{(l, i)} = h^{(l, i)}_T$ or $h^{(l, i)} = \frac{1}{T} \sum_t h^{(l, i)}_t$ where $t \in \{1, \dots, T\}$ denotes the position of the token.
    \item Based on the obtained sufficient statistics, we train two separate Bayesian Ridge models, one for Cor samples and one for Incor ones, assuming an affine dependence of each neuron on the activations of the previous layer $h^{(l, :)}$,
       \begin{align*}
       & y^{(l, i)}  = w_{u, 0}^{(l, i)} + w_u^{(l, i)} \cdot h^{(l-1, :)} + \varepsilon_u^{(l, i)}, \quad \quad \textrm{s.t.} \quad  \varepsilon_u^{(l, i)} \sim \mathcal{N}(0, {\sigma^2}_u^{(l, i)}) \\ 
             & \textrm{with} \quad         w_u^{(l, i)} \sim \mathcal{N}(0, {\lambda^{-1}}_u^{(l, i)} I).
    \end{align*}
    Such Ridge models correspond to Maximum a Posteriori estimation problems and can be solved by the celebrated Expectation Maximization algorithm \cite{watanabe2003algorithm}. However, to remove the dependence on negligible activations which might introduce spurious effects, we reduce the dimensionality of the design matrices via singular value decomposition, effectively performing a Bayesian principal component regression \cite{agarwal2019robustness}.  
        We optimize the choice of the hyperparameters $\lambda$ and $\sigma^2$ with the marginal likelihood iterative procedure from \cite{mackay1992bayesian}.
    \item Based on the layer-level posterior distributions, we obtain a global UQ measure for the LLM via a sparse linear regression predicting the correctness of the LLM. More precisely, we propose two approaches for UQ feature design:  \begin{itemize}
        \item First, we use as features the posterior log-likelihoods: $- \log (p(y^{(l, i)}|h^{(l-1, :)}, u))$, which can be interpreted as the information content of the activation $ y^{(l, i)}$ under $\{u=0\}$ or $\{u=1\}$.
        \item Second, we borrow a heuristic from statistical hypothesis testing that consists in taking the log of the posterior likelihood ratio to determine which option is the most likely between $\{u=0\}$ and $\{u=1\}$:
        \begin{align*}
            \log \left (\frac{p(y^{(l, i)}|h^{(l-1, :)}, u=1)}{p(y^{(l, i)}|h^{(l-1, :)}, u=0)}\right).
        \end{align*}
    \end{itemize}
\end{enumerate}

\section{Experimental Results}

\subsection{Experimental Setup}

\paragraph{Models and Dataset}
Our experiments use the instruction-tuned LLMs Llama-3.1-8B-Instruct \cite{dubey2024llama}, Qwen3-8B \cite{yang2025qwen3}, Ministral-8B-Instruct-2410 \cite{ministral8b2025} and SmolLM3-3B \cite{smollm3b2025}.
The multiple-choice question dataset is the MMLU dataset in the 0-shot setting \cite{hendrycks2020measuring}, using the token with maximum softmax probability for accuracy evaluation. MMLU is a general knowledge dataset extensively used to benchamrk LLM capabilities. Prompts follow the official recommendations if applicable (see Appendix \ref{sec:prompts}). The Bayesian models are trained on the auxiliary training set of MMLU (99,842 examples), while the test set (about 14,042 examples) is reserved for evaluation.

\paragraph{Baselines}
We compare our approach with two baseline methods: (i) the maximum softmax probability (MSP) baseline, which consists in taking the softmax probability of the generated token and represents the current state-of-the-art \cite{Vashurin_2025}, and (ii) the raw hidden states baseline, which consists in taking the raw hidden states instead of the features from the Ridge models as features for the sparse regression. Some works indeed show that raw hidden states contain relevant information for uncertainty quantification \cite{su2024unsupervised, chen2024inside}. We evaluate the latter baseline in two settings: taking only the hidden states of the generated token (A), and averaging the hidden states over the question tokens with the generated token (Q+A).

\paragraph{Metrics and Calibration}
We report the AUROC to measure the quality of the uncertainty estimates. Detailed results are available in Appendix \ref{sec:additional}, with Expected Calibration Error (ECE) included for completeness, as uncertainty scores should be calibrated for practical use. We include ECE pre- and post-calibration after applying isotonic regression.

\subsection{Results}

\begin{table}[H]
\caption{AUROC for the Bayesian Linear Lens combined with maximum softmax probability baseline. Mean and standard deviation over 5 folds are reported. For each LLM, the best value is in bold and the second best is underlined. We mark with a star when the value significantly outperforms the maximum softmax probability baseline. Additional information on the derivation of these scores can be found in the Appendix \ref{sec:implementation}.}
\label{tab:ridge}
\begin{tabular}{lllll}
\toprule
 & Llama-3.1-8B & Qwen3-8B & Ministral-8B & SmolLM3-3B \\
\midrule
MSP & 0.8089 & 0.6498 & 0.7905 & 0.7430 \\
Raw neurons (A) & 0.8112 (0.0020)* & \underline{0.8289} (0.0015)* & 0.7947 (0.0102) & 0.7694 (0.0065)* \\
Raw neurons (Q+A) & 0.8026 (0.0019) & 0.7567 (0.0114)* & 0.7904 (0.0105) & 0.7632 (0.0063)* \\
Ridge (A, Cor.) & 0.8043 (0.0019) & 0.8103 (0.0046)* & 0.7928 (0.0084) & 0.7652 (0.0053)* \\
Ridge (A, Incor.) & 0.8097 (0.0023) & 0.8144 (0.0034)* & 0.7936 (0.0105) & 0.7686 (0.0084)* \\
Ridge (A, Ratio) & \underline{0.8119} (0.0029)* & \textbf{0.8291} (0.0029)* & 0.7948 (0.0093) & \textbf{0.7700} (0.0061)* \\
Ridge (Q+A, Cor.) & 0.8111 (0.0024) & 0.8284 (0.0026)* & \underline{0.7952} (0.0104) & 0.7688 (0.0071)* \\
Ridge (Q+A, Incor.) & 0.8115 (0.0019)* & 0.8280 (0.0023)* & 0.7948 (0.0103) & 0.7693 (0.0068)* \\
Ridge (Q+A, Ratio) & \textbf{0.8119} (0.0020)* & 0.8269 (0.0021)* & \textbf{0.7956} (0.0110) & \underline{0.7697} (0.0067)* \\
\bottomrule
\end{tabular}
\end{table}

The Bayesian Linear Lens achieve significant improvements for 3 out of the 4 LLMs considered, with the most significant ones for Qwen3-8B and SmolLM3-3B and moderate ones for Llama-3.1-8B-Instruct. For Ministral-8B, the difference is not significant with respect to the MSP baseline. Compared to the raw hidden states baseline, the Bayesian Linear Lens only allow small improvements. We partially investigate why the raw hidden states baseline and the Ridge method seem very correlated in Appendix \ref{sec:comparison}.

\subsection{Extension to Open-Ended Questions}
Building on the framework of \cite{kadavath2022language}, we recast open-ended question uncertainty estimation into a structured setting by prompting the LLM to first generate an answer and then evaluate its own correctness. Based on the model’s internal activations during this process, we predict a confidence score for the produced answer. We compare our approach against state-of-the-art methods for open-ended uncertainty estimation, namely Semantic Entropy (SE) \cite{farquhar2024detecting} and reflexive methods from \cite{kadavath2022language}, hereafter referred to as P(True). To reduce the computational and memory costs of training, we implement our method as a sparse probe on the model activations, denoted Probe. Our results show a substantial improvement in the accuracy of confidence estimation, consistent with trends observed in the structured setting.
\begin{table}[H]
\caption{AUROC for the different methods on the TriviaQA dataset. 
For each LLM, the best value is in bold and the second best is underlined. .}
\label{tab:ptrue}
\centering
\begin{tabular}{lllll}
\toprule
 & Llama-3.1-8B & Qwen3-8B & Ministral-8B & SmolLM3-3B \\
\midrule
P(True) & 0.7710 & 0.7442 & 0.8154 & \underline{0.8265} \\
P(True)-10 & 0.7592 & 0.7194 & 0.7823 & 0.8237 \\
SE & \underline{0.8288} & \underline{0.7905} & \underline{0.8352} & 0.8132 \\
Probe (Ours) & \textbf{0.8773}  & \textbf{0.8958} & \textbf{0.8837}  & \textbf{0.8708} \\
\bottomrule
\end{tabular}
\end{table}
\section{Discussion \& Conclusion}

In this work, we proposed a novel and efficient uncertainty estimation approach for LLMs by leveraging activation patterns to detect hallucinated outputs. Targeting factual uncertainty, we designed a lightweight yet principled framework that combines Bayesian posterior likelihoods through sparse linear regression. Despite its simplicity, this approach consistently achieves state-of-the-art performance, surpassing existing methods. These findings highlight that effective uncertainty quantification in LLMs does not necessarily require complex architectures, but can emerge from a rigorous statistical treatment of internal representations. This opens up promising directions for developing scalable, interpretable, and reliable uncertainty-aware LLMs, with immediate applications to safe deployment in automated decision-making systems.
\section*{Broader Impact}
This work advances safe and reliable AI by improving Uncertainty Quantification (UQ) in Large Language Models (LLMs)—a core component of trustworthy deployment alongside robustness \cite{dakhmouche2025noise, rezkellah2025machine, dakhmouche2025network, dakhmoucherobust}, privacy \cite{dong2022gaussian, sajadmanesh2023privacy, sajadmanesh2024progap} and fairness \cite{castro2025representative, zhang2024data}. By offering an efficient alternative to computationally expensive ensemble or deep Bayesian methods, this work contributes to democratizing access to robust AI safety tools while fostering responsible human–AI interaction and reducing overconfidence in automated systems.
\section*{Acknowledgements }
This work was supported by the Swiss National Science Foundation under grant No. 212876. We acknowledge computational resources from the Swiss National Supercomputing Centre CSCS. R.D. and A.L. acknowledge Dr. Ivan Lunati for providing laboratory infrastructure and computational resources. 
\bibliography{Biblio}

\begin{thebibliography}{10}

\bibitem{agarwal2019robustness}
Anish Agarwal, Devavrat Shah, Dennis Shen, and Dogyoon Song.
\newblock On robustness of principal component regression.
\newblock {\em Advances in Neural Information Processing Systems}, 32, 2019.

\bibitem{belrose2023eliciting}
Nora Belrose, Zach Furman, Logan Smith, Danny Halawi, Igor Ostrovsky, Lev McKinney, Stella Biderman, and Jacob Steinhardt.
\newblock Eliciting latent predictions from transformers with the tuned lens.
\newblock {\em arXiv preprint arXiv:2303.08112}, 2023.

\bibitem{castro2025representative}
Clinton Castro and Michele Loi.
\newblock The representative individuals approach to fair machine learning.
\newblock {\em AI and Ethics}, pages 1--11, 2025.

\bibitem{chen2024inside}
Chao Chen, Kai Liu, Ze~Chen, Yi~Gu, Yue Wu, Mingyuan Tao, Zhihang Fu, and Jieping Ye.
\newblock Inside: Llms' internal states retain the power of hallucination detection.
\newblock In {\em The Twelfth International Conference on Learning Representations}.

\bibitem{dakhmouche2025network}
Ramzi Dakhmouche, Ivan Lunati, and Hossein Gorji.
\newblock Network system forecasting despite topology perturbation.
\newblock In {\em ICML 2025 Workshop on Scaling Up Intervention Models}, 2025.

\bibitem{dakhmouche2025noise}
Ramzi Dakhmouche, Ivan Lunati, and Hossein Gorji.
\newblock Noise tolerance of distributionally robust learning.
\newblock In {\em ICML 2025 Workshop on Scaling Up Intervention Models}, 2025.

\bibitem{dakhmoucherobust}
Ramzi Dakhmouche, Ivan Lunati, and Hossein Gorji.
\newblock Robust symbolic regression for dynamical system identification.
\newblock {\em Transactions on Machine Learning Research}, 2025.

\bibitem{dar2023analyzing}
Guy Dar, Mor Geva, Ankit Gupta, and Jonathan Berant.
\newblock Analyzing transformers in embedding space.
\newblock In {\em Proceedings of the 61st Annual Meeting of the Association for Computational Linguistics (Volume 1: Long Papers)}, pages 16124--16170, 2023.

\bibitem{dong2022gaussian}
Jinshuo Dong, Aaron Roth, and Weijie~J Su.
\newblock Gaussian differential privacy.
\newblock {\em Journal of the Royal Statistical Society Series B: Statistical Methodology}, 84(1):3--37, 2022.

\bibitem{duan2024shifting}
Jinhao Duan, Hao Cheng, Shiqi Wang, Alex Zavalny, Chenan Wang, Renjing Xu, Bhavya Kailkhura, and Kaidi Xu.
\newblock Shifting attention to relevance: Towards the predictive uncertainty quantification of free-form large language models.
\newblock In {\em Proceedings of the 62nd Annual Meeting of the Association for Computational Linguistics (Volume 1: Long Papers)}, pages 5050--5063, 2024.

\bibitem{dubey2024llama}
Abhimanyu Dubey, Abhinav Jauhri, Abhinav Pandey, Abhishek Kadian, Ahmad Al-Dahle, Aiesha Letman, Akhil Mathur, Alan Schelten, Amy Yang, Angela Fan, et~al.
\newblock The llama 3 herd of models.
\newblock {\em arXiv e-prints}, pages arXiv--2407, 2024.

\bibitem{farquhar2024detecting}
Sebastian Farquhar, Jannik Kossen, Lorenz Kuhn, and Yarin Gal.
\newblock Detecting hallucinations in large language models using semantic entropy.
\newblock {\em Nature}, 630(8017):625--630, 2024.

\bibitem{halawioverthinking}
Danny Halawi, Jean-Stanislas Denain, and Jacob Steinhardt.
\newblock Overthinking the truth: Understanding how language models process false demonstrations.
\newblock In {\em The Twelfth International Conference on Learning Representations}, 2024.

\bibitem{hendrycks2020measuring}
Dan Hendrycks, Collin Burns, Steven Basart, Andy Zou, Mantas Mazeika, Dawn Song, and Jacob Steinhardt.
\newblock Measuring massive multitask language understanding.
\newblock In {\em International Conference on Learning Representations}.

\bibitem{smollm3b2025}
{HuggingFace}.
\newblock Smollm3-3b.
\newblock \url{https://huggingface.co/HuggingFaceTB/SmolLM3-3B}, 2025.
\newblock Accessed: 2025-09-05.

\bibitem{kadavath2022language}
Saurav Kadavath, Tom Conerly, Amanda Askell, Tom Henighan, Dawn Drain, Ethan Perez, Nicholas Schiefer, Zac Hatfield-Dodds, Nova DasSarma, Eli Tran-Johnson, et~al.
\newblock Language models (mostly) know what they know.
\newblock {\em arXiv preprint arXiv:2207.05221}, 2022.

\bibitem{kendall2017uncertainties}
Alex Kendall and Yarin Gal.
\newblock What uncertainties do we need in bayesian deep learning for computer vision?
\newblock {\em Advances in neural information processing systems}, 30, 2017.

\bibitem{kuhn2023semantic}
Lorenz Kuhn, Yarin Gal, and Sebastian Farquhar.
\newblock Semantic uncertainty: Linguistic invariances for uncertainty estimation in natural language generation.
\newblock In {\em The Eleventh International Conference on Learning Representations}.

\bibitem{lin2024generating}
Zhen Lin, Shubhendu Trivedi, and Jimeng Sun.
\newblock Generating with confidence: Uncertainty quantification for black-box large language models.
\newblock {\em Transactions on Machine Learning Research}, 2024.

\bibitem{lindsey2025biology}
Jack Lindsey, Wes Gurnee, Emmanuel Ameisen, Brian Chen, Adam Pearce, Nicholas~L. Turner, Craig Citro, David Abrahams, Shan Carter, Basil Hosmer, Jonathan Marcus, Michael Sklar, Adly Templeton, Trenton Bricken, Callum McDougall, Hoagy Cunningham, Thomas Henighan, Adam Jermyn, Andy Jones, Andrew Persic, Zhenyi Qi, T.~Ben Thompson, Sam Zimmerman, Kelley Rivoire, Thomas Conerly, Chris Olah, and Joshua Batson.
\newblock On the biology of a large language model.
\newblock {\em Transformer Circuits Thread}, 2025.

\bibitem{mackay1992bayesian}
David~JC MacKay.
\newblock Bayesian interpolation.
\newblock {\em Neural computation}, 4(3):415--447, 1992.

\bibitem{ministral8b2025}
{Mistral AI}.
\newblock Ministral-8b-instruct-2410.
\newblock \url{https://huggingface.co/mistralai/Ministral-8B-Instruct-2410}, 2025.
\newblock Accessed: 2025-09-05.

\bibitem{scikit-learn}
Fabian Pedregosa, Ga\"el Varoquaux, Alexandre Gramfort, Vincent Michel, Bertrand Thirion, Olivier Grisel, Mathieu Blondel, Peter Prettenhofer, Ron Weiss, Vincent Dubourg, Jake VanderPlas, Alexandre Passos, David Cournapeau, Matthieu Brucher, Matthieu Perrot, and Édouard Duchesnay.
\newblock Scikit-learn: Machine learning in {Python}.
\newblock {\em Journal of Machine Learning Research}, 12:2825--2830, 2011.

\bibitem{rezkellah2025machine}
Fatmazohra Rezkellah and Ramzi Dakhmouche.
\newblock Machine unlearning meets adversarial robustness via constrained interventions on llms.
\newblock {\em arXiv preprint arXiv:2510.03567}, 2025.

\bibitem{sajadmanesh2023privacy}
Sina Sajadmanesh.
\newblock Privacy-preserving machine learning on graphs.
\newblock Technical report, Technical Report. EPFL, 2023.

\bibitem{sajadmanesh2024progap}
Sina Sajadmanesh and Daniel Gatica-Perez.
\newblock Progap: Progressive graph neural networks with differential privacy guarantees.
\newblock In {\em Proceedings of the 17th ACM International Conference on Web Search and Data Mining}, pages 596--605, 2024.

\bibitem{su2024unsupervised}
Weihang Su, Changyue Wang, Qingyao Ai, Yiran Hu, Zhijing Wu, Yujia Zhou, and Yiqun Liu.
\newblock Unsupervised real-time hallucination detection based on the internal states of large language models.
\newblock In {\em Findings of the Association for Computational Linguistics ACL 2024}, pages 14379--14391, 2024.

\bibitem{Vashurin_2025}
Roman Vashurin, Ekaterina Fadeeva, Artem Vazhentsev, Lyudmila Rvanova, Daniil Vasilev, Akim Tsvigun, Sergey Petrakov, Rui Xing, Abdelrahman Sadallah, Kirill Grishchenkov, Alexander Panchenko, Timothy Baldwin, Preslav Nakov, Maxim Panov, and Artem Shelmanov.
\newblock Benchmarking uncertainty quantification methods for large language models with lm-polygraph.
\newblock {\em Transactions of the Association for Computational Linguistics}, 13:220–248, 2025.

\bibitem{watanabe2003algorithm}
Michiko Watanabe and Kazunori Yamaguchi.
\newblock {\em The EM algorithm and related statistical models}.
\newblock CRC Press, 2003.

\bibitem{yang2025qwen3}
An~Yang, Anfeng Li, Baosong Yang, Beichen Zhang, Binyuan Hui, Bo~Zheng, Bowen Yu, Chang Gao, Chengen Huang, Chenxu Lv, et~al.
\newblock Qwen3 technical report.
\newblock {\em arXiv preprint arXiv:2505.09388}, 2025.

\bibitem{zhang2024data}
Haoran Zhang, Walter Gerych, and Marzyeh Ghassemi.
\newblock A data-centric perspective to fair machine learning for healthcare.
\newblock {\em Nature Reviews Methods Primers}, 4(1):86, 2024.

\end{thebibliography}
\bibliographystyle{plain}

\newpage
\appendix

\section{Additional Numerical Results}
\label{sec:additional}

For the purpose of completeness, we also consider non-Bayesian methods in \ref{sec:bayesian} for training the probes. Namely, we experiment with: \begin{itemize}
    \item simple Gaussian density. The corresponding model is simply:
    \begin{align*}
        y^{(l, i)}|u=0 \sim \mathcal{N}(\mu_0^{(l, i)}, {\sigma^2}_0^{(l, i)}) \\
        y^{(l, i)}|u=1 \sim \mathcal{N}(\mu_1^{(l, i)}, {\sigma^2}_1^{(l, i)})            
    \end{align*}
        
    \item simple Gaussian regression over the SVD-truncated previous hidden states. From a modeling perspective, it makes sense to model a residual neural network as Markovian, where the next hidden state depends only on the previous hidden state. Formally, we can assume that $p(h^{(l, i)} | (h^{(m, i))})_{m \leq l-1, i \in D}) = p(h^{(l, i)} | (h^{(l-1, i))})_{i \in D})$. The corresponding linear model under $\{u=0\}$ is:
    \begin{align*}
        y^{(l, i)} = w_{0, 0}^{(l, i)} + w_0^{(l, i)} \cdot h_K^{(l-1, :)} + \varepsilon_0^{(l, i)}
    \end{align*}

    and under $\{u=1\}$:
    \begin{align*}
        y^{(l, i)} = w_{1, 0}^{(l, i)} + w_1^{(l, i)} \cdot h_K^{(l-1, :)} + \varepsilon_1^{(l, i)}
    \end{align*}

    where $h_K^{(l-1, :)}$ is the $K-$dimensional vector of the truncated previous hidden state and with the Gaussian noise assumption:
   \begin{align*}
        \varepsilon_0^{(l, i)} \sim \mathcal{N}(0, {\sigma^2}_0^{(l, i)}) \\
        \varepsilon_1^{(l, i)} \sim \mathcal{N}(0, {\sigma^2}_1^{(l, i)})
    \end{align*}
\end{itemize}

As in the Bayesian case, we combine the neuron-level features to obtain uncertainty scores by performing Elastic-Net logistic regression.

Based on the layer-level posterior distributions, we obtain a global UQ measure for the LLM via a sparse linear regression predicting the correctness of the LLM. More precisely, we propose two approaches for UQ feature design:  \begin{itemize}
        \item First, we use as features the posterior log-likelihoods: $- \log (p(y^{(l, i)}|u))$ for the simple density model and $- \log (p(y^{(l, i)}|h_K^{(l-1, :)}, u))$ for the regression model, which can be interpreted as the information content of the activation $ y^{(l, i)}$ under $\{u=0\}$ or $\{u=1\}$.
        \item Second, we borrow a heuristic from statistical hypothesis testing that consists in taking the log of the posterior likelihood ratio to compare which option is the most likely between $\{u=0\}$ and $\{u=1\}$. For the simple density model:
    \begin{align*}
        \log(\frac{p(y^{(l, i)}|u=1)}{p(y^{(l, i)}|u=0)})
    \end{align*}        
    For the regression model:
    \begin{align*}
        \log(\frac{p(y^{(l, i)}|h_K^{(l-1, :)}, u=1)}{p(y^{(l, i)}|h_K^{(l-1, :)}, u=0)})
    \end{align*}
\end{itemize}

Surprisingly, we find that even these non-Bayesian methods are also able to achieve significantly better results than the MSP baseline. 

\begin{table}[H]
\caption{AUROC for various methods combined with maximum softmax probability baseline. Mean and standard deviation over 5 folds are reported. For each LLM, the best value is in bold and the second best is underlined. We mark with a star when the value significantly outperforms the maximum softmax probability baseline.}
\label{tab:example}
\begin{tabular}{lllll}
\toprule
 & Llama-3.1-8B & Qwen3-8B & Ministral-8B & SmolLM3-3B \\
\midrule
MSP & 0.8089 & 0.6498 & 0.7905 & 0.7430 \\
Raw neurons (A) & 0.8112 (0.0020)* & 0.8289 (0.0015)* & 0.7947 (0.0102) & 0.7694 (0.0065)* \\
Raw neurons (Q+A) & 0.8026 (0.0019) & 0.7567 (0.0114)* & 0.7904 (0.0105) & 0.7632 (0.0063)* \\
Density (A, Cor.) & 0.8071 (0.0015) & 0.8140 (0.0020)* & 0.7913 (0.0104) & 0.7701 (0.0058)* \\
Density (A, Incor.) & 0.8098 (0.0015) & 0.8186 (0.0044)* & 0.7935 (0.0103) & 0.7699 (0.0054)* \\
Density (A, Ratio) & \underline{0.8121} (0.0016)* & 0.8289 (0.0022)* & 0.7964 (0.0106) & \textbf{0.7725} (0.0063)* \\
Density (Q+A, Cor.) & 0.8110 (0.0012)* & 0.8272 (0.0036)* & \textbf{0.7966} (0.0096) & 0.7715 (0.0063)* \\
Density (Q+A, Incor.) & 0.8111 (0.0011)* & 0.8272 (0.0035)* & \underline{0.7965} (0.0097) & 0.7716 (0.0063)* \\
Density (Q+A, Ratio) & 0.8107 (0.0018)* & \underline{0.8292} (0.0036)* & 0.7960 (0.0098) & 0.7705 (0.0064)* \\
Regression (A, Cor.) & 0.8043 (0.0019) & 0.8103 (0.0046)* & 0.7929 (0.0085) & 0.7652 (0.0053)* \\
Regression (A, Incor.) & 0.8097 (0.0024) & 0.8144 (0.0035)* & 0.7937 (0.0105) & 0.7685 (0.0082)* \\
Regression (A, Ratio) & \textbf{0.8124} (0.0019)* & \textbf{0.8297} (0.0019)* & 0.7943 (0.0096) & \underline{0.7723} (0.0065)* \\
Regression (Q+A, Cor.) & 0.8111 (0.0024) & 0.8284 (0.0026)* & 0.7952 (0.0104) & 0.7688 (0.0071)* \\
Regression (Q+A, Incor.) & 0.8115 (0.0019)* & 0.8281 (0.0023)* & 0.7948 (0.0103) & 0.7693 (0.0068)* \\
Regression (Q+A, Ratio) & 0.8113 (0.0024)* & 0.8288 (0.0034)* & 0.7954 (0.0092) & 0.7699 (0.0068)* \\
\bottomrule
\end{tabular}
\end{table}

Below, we report detailed results for each model, including calibration metrics. Across all models, we also notice that that the log-likelihood ratios perform better than taking separately the accepted (Cor) or rejected (Incor) model. We find that taking only the hidden states of the generated token (A) gives better results than averaging the hidden states over the question tokens and the generated token (Q+A).

\begin{table}[H]
\caption{Llama-3.1-8B-Instruct. Mean and standard deviation over 5 folds are reported. For each metric, the best value is in bold and the second best is underlined. For the AUROC, we mark with a star when the value significantly outperforms the maximum softmax probability baseline.}
\label{tab:llama}
\begin{tabular}{lllll}
\toprule
 & AUC & AUC (combined) & ECE & ECE (calib.) \\
\midrule
Maximum softmax probability & \textbf{0.8089} & 0.8089 & 0.1242 & 0.0223 (0.0040) \\
Raw neurons (A) & 0.8045 (0.0027) & 0.8112 (0.0020)* & 0.0290 (0.0049) & 0.0269 (0.0037) \\
Raw neurons (Q+A) & 0.7182 (0.0028) & 0.8026 (0.0019) & 0.0300 (0.0023) & 0.0296 (0.0026) \\
Density (A, Cor.) & 0.7983 (0.0020) & 0.8071 (0.0015) & 0.0545 (0.0064) & \underline{0.0218} (0.0020) \\
Density (A, Incor.) & 0.8005 (0.0025) & 0.8098 (0.0015) & 0.0287 (0.0012) & 0.0233 (0.0042) \\
Density (A, Ratio) & 0.8074 (0.0026) & \underline{0.8121} (0.0016)* & 0.0254 (0.0030) & \textbf{0.0208} (0.0016) \\
Density (Q+A, Cor.) & 0.8052 (0.0019) & 0.8110 (0.0012)* & \textbf{0.0241} (0.0014) & 0.0229 (0.0049) \\
Density (Q+A, Incor.) & 0.8053 (0.0019) & 0.8111 (0.0011)* & \underline{0.0244} (0.0035) & 0.0227 (0.0025) \\
Density (Q+A, Ratio) & 0.8049 (0.0024) & 0.8107 (0.0018)* & 0.0274 (0.0014) & 0.0225 (0.0050) \\
Truncated Regression (A, Cor.) & 0.7728 (0.0050) & 0.8043 (0.0019) & 0.0437 (0.0018) & 0.0244 (0.0086) \\
Truncated Regression (A, Incor.) & 0.7873 (0.0025) & 0.8097 (0.0024) & 0.0290 (0.0015) & 0.0262 (0.0024) \\
Truncated Regression (A, Ratio) & 0.8070 (0.0026) & \textbf{0.8124} (0.0019)* & 0.0307 (0.0026) & 0.0282 (0.0037) \\
Truncated Regression (Q+A, Cor.) & 0.8049 (0.0023) & 0.8111 (0.0024) & 0.0305 (0.0044) & 0.0255 (0.0045) \\
Truncated Regression (Q+A, Incor.) & 0.8055 (0.0018) & 0.8115 (0.0019)* & 0.0266 (0.0025) & 0.0244 (0.0031) \\
Truncated Regression (Q+A, Ratio) & 0.8059 (0.0025) & 0.8113 (0.0024)* & 0.0285 (0.0030) & 0.0225 (0.0027) \\
Ridge (A, Cor.) & 0.7728 (0.0050) & 0.8043 (0.0019) & 0.0433 (0.0021) & 0.0237 (0.0088) \\
Ridge (A, Incor.) & 0.7872 (0.0025) & 0.8097 (0.0023) & 0.0269 (0.0026) & 0.0242 (0.0016) \\
Ridge (A, Ratio) & 0.8070 (0.0034) & 0.8119 (0.0029)* & 0.0299 (0.0025) & 0.0270 (0.0039) \\
Ridge (Q+A, Cor.) & 0.8050 (0.0023) & 0.8111 (0.0024) & 0.0313 (0.0041) & 0.0273 (0.0031) \\
Ridge (Q+A, Incor.) & 0.8054 (0.0019) & 0.8115 (0.0019)* & 0.0250 (0.0023) & 0.0232 (0.0040) \\
Ridge (Q+A, Ratio) & \underline{0.8075} (0.0020) & 0.8119 (0.0020)* & 0.0283 (0.0054) & 0.0289 (0.0039) \\
\bottomrule
\end{tabular}
\end{table}

\begin{table}[H]
\caption{Qwen3-8B. Mean and standard deviation over 5 folds are reported. For each metric, the best value is in bold and the second best is underlined. For the AUROC, we mark with a star when the value significantly outperforms the maximum softmax probability baseline.}
\label{tab:qwen}
\begin{tabular}{lllll}
\toprule
 & AUC & AUC (combined) & ECE & ECE (calib.) \\
\midrule
Maximum softmax probability & 0.6498 & 0.6498 & 0.2580 & \textbf{0.0132} (0.0050) \\
Raw neurons (A) & 0.8269 (0.0020)* & 0.8289 (0.0015)* & 0.0331 (0.0057) & 0.0255 (0.0033) \\
Raw neurons (Q+A) & 0.7313 (0.0126)* & 0.7567 (0.0114)* & 0.0299 (0.0066) & 0.0291 (0.0062) \\
Density (A, Cor.) & 0.8119 (0.0023)* & 0.8140 (0.0020)* & 0.0498 (0.0108) & 0.0274 (0.0057) \\
Density (A, Incor.) & 0.8159 (0.0044)* & 0.8186 (0.0044)* & 0.0255 (0.0043) & 0.0238 (0.0053) \\
Density (A, Ratio) & 0.8274 (0.0023)* & 0.8289 (0.0022)* & 0.0282 (0.0034) & 0.0272 (0.0031) \\
Density (Q+A, Cor.) & 0.8256 (0.0037)* & 0.8272 (0.0036)* & 0.0277 (0.0045) & 0.0275 (0.0042) \\
Density (Q+A, Incor.) & 0.8257 (0.0035)* & 0.8272 (0.0035)* & 0.0267 (0.0052) & 0.0248 (0.0043) \\
Density (Q+A, Ratio) & \underline{0.8276} (0.0036)* & \underline{0.8292} (0.0036)* & 0.0287 (0.0055) & 0.0256 (0.0065) \\
Truncated Regression (A, Cor.) & 0.8056 (0.0064)* & 0.8103 (0.0046)* & 0.0474 (0.0100) & 0.0279 (0.0078) \\
Truncated Regression (A, Incor.) & 0.8083 (0.0046)* & 0.8144 (0.0035)* & 0.0310 (0.0042) & 0.0291 (0.0049) \\
Truncated Regression (A, Ratio) & \textbf{0.8280} (0.0020)* & \textbf{0.8297} (0.0019)* & 0.0291 (0.0041) & 0.0253 (0.0041) \\
Truncated Regression (Q+A, Cor.) & 0.8270 (0.0026)* & 0.8284 (0.0026)* & \textbf{0.0247} (0.0042) & \underline{0.0220} (0.0046) \\
Truncated Regression (Q+A, Incor.) & 0.8266 (0.0025)* & 0.8281 (0.0023)* & 0.0259 (0.0036) & 0.0256 (0.0028) \\
Truncated Regression (Q+A, Ratio) & 0.8271 (0.0036)* & 0.8288 (0.0034)* & 0.0284 (0.0046) & 0.0235 (0.0063) \\
Ridge (A, Cor.) & 0.8056 (0.0064)* & 0.8103 (0.0046)* & 0.0472 (0.0101) & 0.0284 (0.0084) \\
Ridge (A, Incor.) & 0.8084 (0.0046)* & 0.8144 (0.0034)* & 0.0296 (0.0049) & 0.0310 (0.0065) \\
Ridge (A, Ratio) & 0.8273 (0.0023)* & 0.8291 (0.0029)* & 0.0306 (0.0060) & 0.0228 (0.0034) \\
Ridge (Q+A, Cor.) & 0.8270 (0.0026)* & 0.8284 (0.0026)* & \underline{0.0247} (0.0043) & 0.0222 (0.0047) \\
Ridge (Q+A, Incor.) & 0.8265 (0.0024)* & 0.8280 (0.0023)* & 0.0261 (0.0042) & 0.0267 (0.0029) \\
Ridge (Q+A, Ratio) & 0.8253 (0.0025)* & 0.8269 (0.0021)* & 0.0278 (0.0025) & 0.0231 (0.0026) \\
\bottomrule
\end{tabular}
\end{table}

\begin{table}[H]
\caption{Ministral-8B-Instruct-2410. Mean and standard deviation over 5 folds are reported. For each metric, the best value is in bold and the second best is underlined. For the AUROC, we mark with a star when the value significantly outperforms the maximum softmax probability baseline.}
\label{tab:mistral}
\begin{tabular}{lllll}
\toprule
 & AUC & AUC (combined) & ECE & ECE (calib.) \\
\midrule
Maximum softmax probability & 0.7905 & 0.7905 & 0.0930 & \textbf{0.0240} (0.0049) \\
Raw neurons (A) & 0.7888 (0.0108) & 0.7947 (0.0102) & 0.0353 (0.0035) & 0.0310 (0.0062) \\
Raw neurons (Q+A) & 0.7125 (0.0102) & 0.7904 (0.0105) & 0.0323 (0.0033) & 0.0314 (0.0083) \\
Density (A, Cor.) & 0.7821 (0.0106) & 0.7913 (0.0104) & 0.0382 (0.0079) & 0.0263 (0.0040) \\
Density (A, Incor.) & 0.7857 (0.0109) & 0.7935 (0.0103) & 0.0322 (0.0051) & 0.0299 (0.0087) \\
Density (A, Ratio) & \textbf{0.7915} (0.0106) & 0.7964 (0.0106) & \textbf{0.0280} (0.0046) & \underline{0.0253} (0.0072) \\
Density (Q+A, Cor.) & 0.7910 (0.0096) & \textbf{0.7966} (0.0096) & \underline{0.0284} (0.0049) & 0.0283 (0.0043) \\
Density (Q+A, Incor.) & 0.7910 (0.0097) & \underline{0.7965} (0.0097) & 0.0303 (0.0038) & 0.0283 (0.0041) \\
Density (Q+A, Ratio) & \underline{0.7912} (0.0090) & 0.7960 (0.0098) & 0.0318 (0.0041) & 0.0306 (0.0052) \\
Truncated Regression (A, Cor.) & 0.7636 (0.0083) & 0.7929 (0.0085) & 0.0360 (0.0053) & 0.0281 (0.0073) \\
Truncated Regression (A, Incor.) & 0.7796 (0.0096) & 0.7937 (0.0105) & 0.0327 (0.0058) & 0.0306 (0.0064) \\
Truncated Regression (A, Ratio) & 0.7892 (0.0103) & 0.7943 (0.0096) & 0.0304 (0.0028) & 0.0255 (0.0047) \\
Truncated Regression (Q+A, Cor.) & 0.7903 (0.0100) & 0.7952 (0.0104) & 0.0293 (0.0050) & 0.0278 (0.0032) \\
Truncated Regression (Q+A, Incor.) & 0.7897 (0.0098) & 0.7948 (0.0103) & 0.0295 (0.0049) & 0.0280 (0.0044) \\
Truncated Regression (Q+A, Ratio) & 0.7893 (0.0090) & 0.7954 (0.0092) & 0.0312 (0.0054) & 0.0298 (0.0050) \\
Ridge (A, Cor.) & 0.7635 (0.0081) & 0.7928 (0.0084) & 0.0363 (0.0046) & 0.0304 (0.0050) \\
Ridge (A, Incor.) & 0.7797 (0.0095) & 0.7936 (0.0105) & 0.0325 (0.0066) & 0.0287 (0.0050) \\
Ridge (A, Ratio) & 0.7895 (0.0096) & 0.7948 (0.0093) & 0.0327 (0.0035) & 0.0261 (0.0073) \\
Ridge (Q+A, Cor.) & 0.7903 (0.0100) & 0.7952 (0.0104) & 0.0293 (0.0050) & 0.0275 (0.0031) \\
Ridge (Q+A, Incor.) & 0.7897 (0.0098) & 0.7948 (0.0103) & 0.0295 (0.0049) & 0.0279 (0.0042) \\
Ridge (Q+A, Ratio) & 0.7899 (0.0110) & 0.7956 (0.0110) & 0.0304 (0.0055) & 0.0259 (0.0089) \\
\bottomrule
\end{tabular}
\end{table}

\begin{table}[H]
\caption{SmolLM3-3B. Mean and standard deviation over 5 folds are reported. For each metric, the best value is in bold and the second best is underlined. For the AUROC, we mark with a star when the value significantly outperforms the maximum softmax probability baseline.}
\label{tab:smollm}
\begin{tabular}{lllll}
\toprule
 & AUC & AUC (combined) & ECE & ECE (calib.) \\
\midrule
Maximum softmax probability & 0.7430 & 0.7430 & 0.1956 & \textbf{0.0209} (0.0033) \\
Raw neurons (A) & 0.7648 (0.0069)* & 0.7694 (0.0065)* & 0.0383 (0.0044) & 0.0278 (0.0058) \\
Raw neurons (Q+A) & 0.6999 (0.0037) & 0.7632 (0.0063)* & 0.0344 (0.0063) & 0.0290 (0.0056) \\
Density (A, Cor.) & 0.7660 (0.0058)* & 0.7701 (0.0058)* & 0.0314 (0.0050) & 0.0255 (0.0037) \\
Density (A, Incor.) & 0.7630 (0.0053)* & 0.7699 (0.0054)* & \textbf{0.0290} (0.0059) & 0.0249 (0.0065) \\
Density (A, Ratio) & \textbf{0.7687} (0.0075)* & \textbf{0.7725} (0.0063)* & \underline{0.0297} (0.0042) & \underline{0.0232} (0.0052) \\
Density (Q+A, Cor.) & 0.7672 (0.0069)* & 0.7715 (0.0063)* & 0.0304 (0.0076) & 0.0270 (0.0050) \\
Density (Q+A, Incor.) & 0.7672 (0.0069)* & 0.7716 (0.0063)* & 0.0311 (0.0075) & 0.0272 (0.0049) \\
Density (Q+A, Ratio) & 0.7659 (0.0069)* & 0.7705 (0.0064)* & 0.0317 (0.0032) & 0.0278 (0.0048) \\
Truncated Regression (A, Cor.) & 0.7469 (0.0060) & 0.7652 (0.0053)* & 0.0405 (0.0077) & 0.0268 (0.0053) \\
Truncated Regression (A, Incor.) & 0.7568 (0.0094)* & 0.7685 (0.0082)* & 0.0325 (0.0061) & 0.0252 (0.0024) \\
Truncated Regression (A, Ratio) & \underline{0.7674} (0.0072)* & \underline{0.7723} (0.0065)* & 0.0359 (0.0029) & 0.0264 (0.0039) \\
Truncated Regression (Q+A, Cor.) & 0.7643 (0.0079)* & 0.7688 (0.0071)* & 0.0335 (0.0044) & 0.0307 (0.0037) \\
Truncated Regression (Q+A, Incor.) & 0.7647 (0.0076)* & 0.7693 (0.0068)* & 0.0315 (0.0047) & 0.0278 (0.0082) \\
Truncated Regression (Q+A, Ratio) & 0.7654 (0.0073)* & 0.7699 (0.0068)* & 0.0333 (0.0019) & 0.0286 (0.0041) \\
Ridge (A, Cor.) & 0.7469 (0.0060) & 0.7652 (0.0053)* & 0.0395 (0.0079) & 0.0246 (0.0027) \\
Ridge (A, Incor.) & 0.7570 (0.0098)* & 0.7686 (0.0084)* & 0.0335 (0.0061) & 0.0273 (0.0028) \\
Ridge (A, Ratio) & 0.7652 (0.0069)* & 0.7700 (0.0061)* & 0.0332 (0.0082) & 0.0251 (0.0035) \\
Ridge (Q+A, Cor.) & 0.7643 (0.0079)* & 0.7688 (0.0071)* & 0.0334 (0.0045) & 0.0307 (0.0036) \\
Ridge (Q+A, Incor.) & 0.7647 (0.0076)* & 0.7693 (0.0068)* & 0.0315 (0.0047) & 0.0277 (0.0078) \\
Ridge (Q+A, Ratio) & 0.7654 (0.0066)* & 0.7697 (0.0067)* & 0.0363 (0.0039) & 0.0247 (0.0025) \\
\bottomrule
\end{tabular}
\end{table}

\newpage
\section{Active Neuron Visualization}
\label{sec:comparison}

\begin{table}[H]
\caption{Correlations of the methods for Llama-3.1-8B-Instruct. Results are given for the answer log-likelihood ratio setting.}
\begin{tabularx}{\textwidth}{@{} *{6}{>{\centering\arraybackslash}X} @{}}
\toprule
 & MSP & Raw neurons & Density & Regression & Ridge \\
\midrule
MSP & 1.0000 & 0.9197 & 0.9256 & 0.9132 & 0.9135 \\
Raw neurons & 0.9197 & 1.0000 & 0.9677 & 0.9647 & 0.9652 \\
Density & 0.9256 & 0.9677 & 1.0000 & 0.9624 & 0.9621 \\
Regression & 0.9132 & 0.9647 & 0.9624 & 1.0000 & 0.9824 \\
Ridge & 0.9135 & 0.9652 & 0.9621 & 0.9824 & 1.0000 \\
\bottomrule
\end{tabularx}
\end{table}

\begin{figure}[H]
    \centering
    \includegraphics[width=\textwidth]{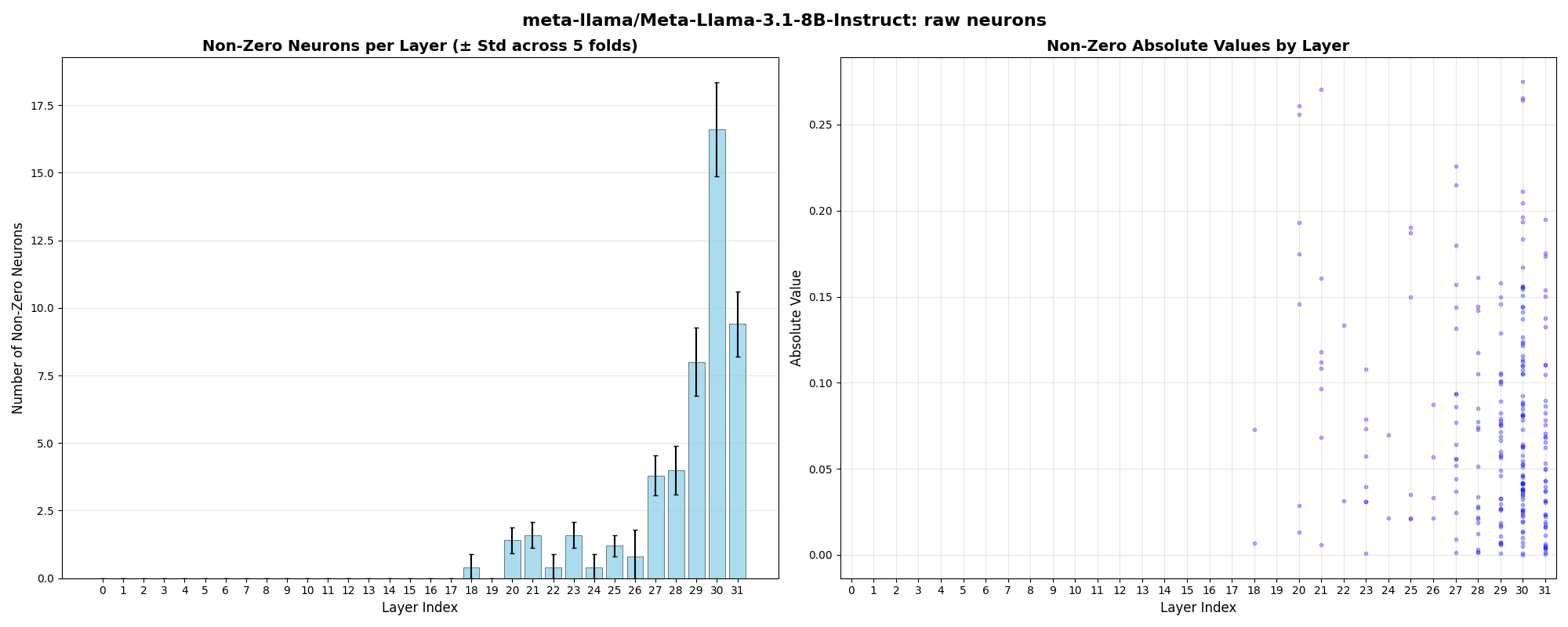}
    \caption{Layers of the activated neurons, Llama-3.1-8B-Instruct, raw neurons, in the answer log-likelihood ratio setting.}
\end{figure}

\begin{figure}[H]
    \centering
    \includegraphics[width=\textwidth]{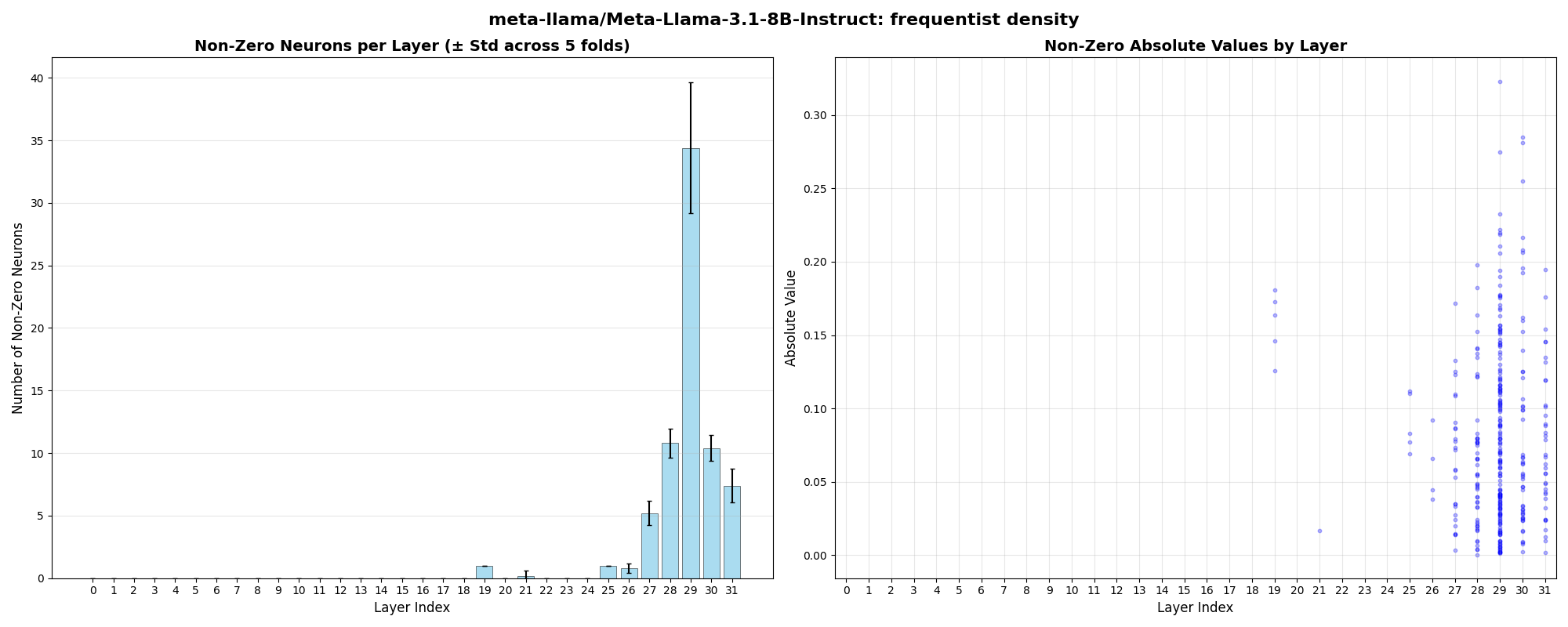}
    \caption{Layers of the activated neurons, Llama-3.1-8B-Instruct, density, in the answer log-likelihood ratio setting.}
\end{figure}

\begin{figure}[H]
    \centering
    \includegraphics[width=\textwidth]{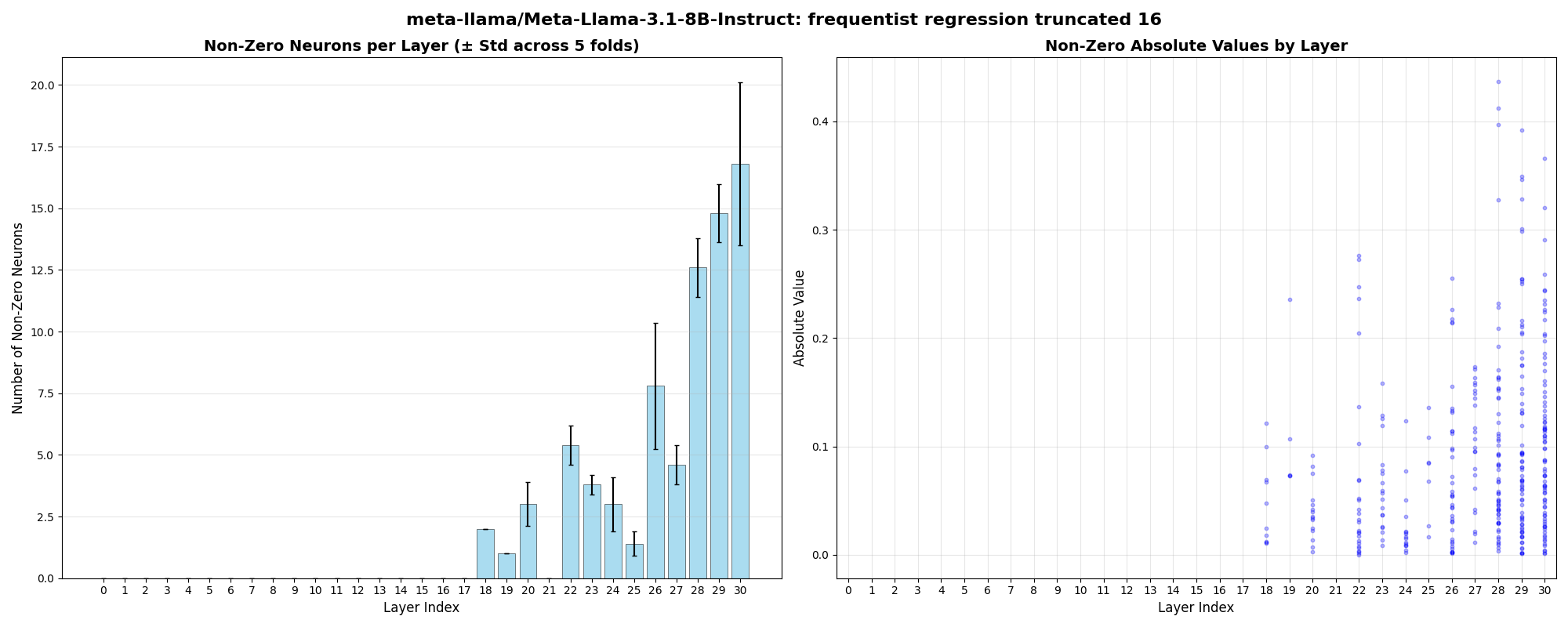}
    \caption{Layers of the activated neurons, Llama-3.1-8B-Instruct, truncated regression, in the answer log-likelihood ratio setting.}
\end{figure}

\begin{figure}[H]
    \centering
    \includegraphics[width=\textwidth]{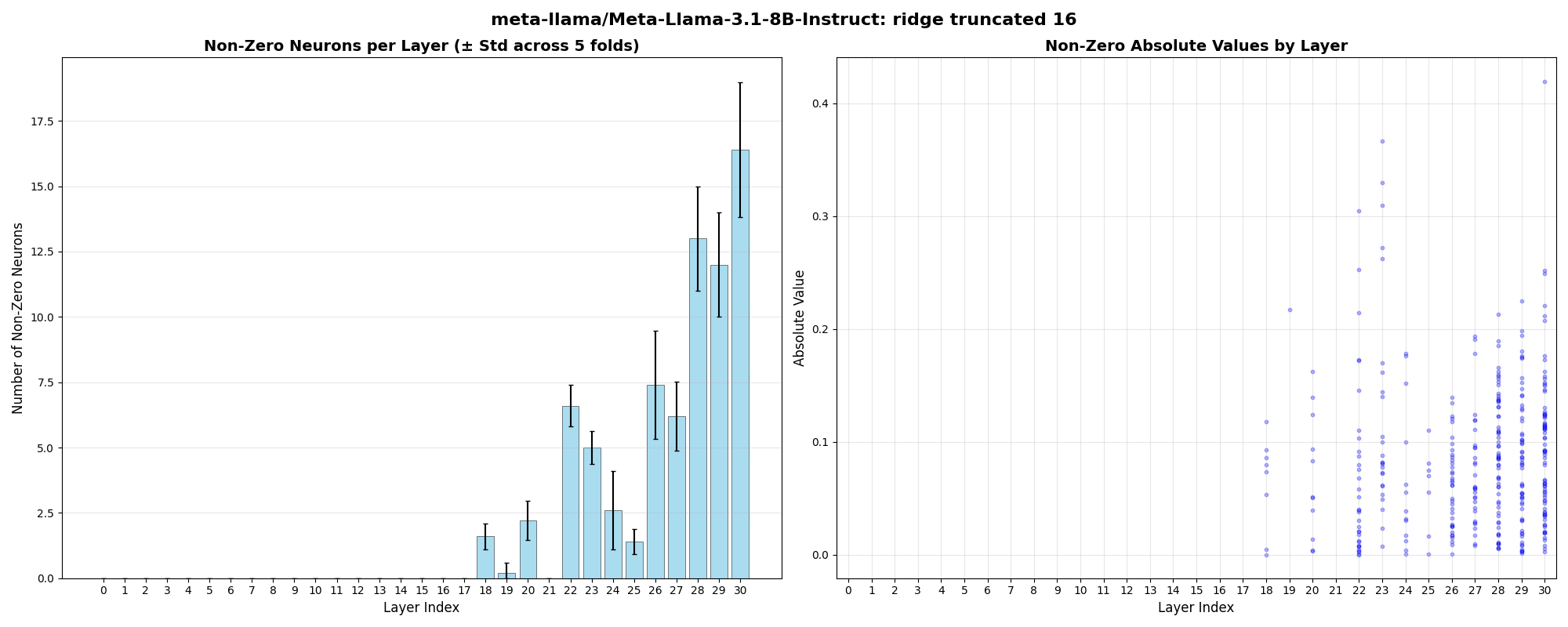}
    \caption{Layers of the activated neurons, Llama-3.1-8B-Instruct, truncated ridge, in the answer log-likelihood ratio setting.}
\end{figure}

\newpage
\section{Prompts}
\label{sec:prompts}

\begin{figure}[H]
\caption{Prompt for Llama-3.1-8B-Instruct, matching the template found on \href{https://huggingface.co/datasets/meta-llama/Llama-3.1-8B-Instruct-evals}{HuggingFace}.}
\begin{lstlisting}[
    basicstyle=\ttfamily,
    breaklines=true,
    numbers=left,
    numberstyle=\tiny,
    frame=single,
    xleftmargin=20pt,
    xrightmargin=20pt,
    resetmargins=true,
    tabsize=4,
    showspaces=false,
    showstringspaces=false,
    showtabs=false,
    gobble=0
]
<|start_header_id|>user<|end_header_id|>

Given the following question and four candidate answers (A, B, C and D), choose the best answer.
Question: {}
A. {}
B. {}
C. {}
D. {}
Your response should end with "The best answer is [the_answer_letter]" where the [the_answer_letter] is one of A, B, C or D.<|eot_id|><|start_header_id|>assistant<|end_header_id|>

The best answer is
\end{lstlisting}
\end{figure}

\begin{figure}[H]
\caption{Prompt for Qwen3-8B, following the recommendations found on \href{https://huggingface.co/Qwen/Qwen3-8B}{HuggingFace}.}
\begin{lstlisting}[
    basicstyle=\ttfamily,
    breaklines=true,
    numbers=left,
    numberstyle=\tiny,
    frame=single,
    xleftmargin=20pt,
    xrightmargin=20pt,
    resetmargins=true,
    tabsize=4,
    showspaces=false,
    showstringspaces=false,
    showtabs=false,
    gobble=0
]
<|im_start|>user
Given the following question and four candidate answers (A, B, C and D), choose the best answer.
Question: {}
A. {}
B. {}
C. {}
D. {}
Start your response by showing your choice in the answer field with only the choice letter, e.g., "answer": "C". /no_think<|im_end|>
<|im_start|>assistant
<think>

</think>

"answer": "
\end{lstlisting}
\end{figure}

\begin{figure}[H]
\caption{Prompt for Ministral-8B-Instruct-2410. No official recommendations were found, so the Llama template was adapted.}
\begin{lstlisting}[
    basicstyle=\ttfamily,
    breaklines=true,
    numbers=left,
    numberstyle=\tiny,
    frame=single,
    xleftmargin=20pt,
    xrightmargin=20pt,
    resetmargins=true,
    tabsize=4,
    showspaces=false,
    showstringspaces=false,
    showtabs=false,
    gobble=0
]
[INST]Given the following question and four candidate answers (A, B, C and D), choose the best answer.
Question: {}
A. {}
B. {}
C. {}
D. {}
Your response should start with "The best answer is [the_answer_letter]" where the [the_answer_letter] is one of A, B, C or D.[/INST]The best answer is
\end{lstlisting}
\end{figure}

\begin{figure}[H]
\caption{Prompt for SmolLM3-3B. No official recommendations were found, so the Llama template was adapted.}
\begin{lstlisting}[
    basicstyle=\ttfamily,
    breaklines=true,
    numbers=left,
    numberstyle=\tiny,
    frame=single,
    xleftmargin=20pt,
    xrightmargin=20pt,
    resetmargins=true,
    tabsize=4,
    showspaces=false,
    showstringspaces=false,
    showtabs=false,
    gobble=0
]
<|im_start|>system
## Metadata

Knowledge Cutoff Date: June 2025
Today Date: 04 September 2025
Reasoning Mode: /no_think

## Custom Instructions

For the question below, choose the best answer from options A, B, C, or D. Your response should start with "Answer: [the_answer_letter]" where the [the_answer_letter] is one of A, B, C or D.

<|im_start|>user
Question: {}
A. {}
B. {}
C. {}
D. {}
<|im_end|>
<|im_start|>assistant
<think>

</think>
Answer:
\end{lstlisting}
\end{figure}

\newpage
\section{Additional implementation details}
\label{sec:implementation}

\paragraph{Ridge regressions hyperparameters} For the truncated regressions we use for simplicity the value $K=16$ for all models, but it could be further tuned. For the hyperparameters of the Ridge regression, we use the marginal likelihood optimization from \cite{mackay1992bayesian}. Our implementation closely follows scikit-learn's BayesianRidge \cite{scikit-learn}.

\paragraph{Post-processing to obtain Uncertainty Estimates}
We store the hidden states, density- and regression-based features during the evaluation on the MMLU test set. These features are then fed to an Elastic-Net regularized logistic regression (predicting the truthfullness of the LLM) to derive the final uncertainty scores.
To obtain final results, we perform nested cross validation: 5 outer loops estimate performance across the dataset, while 4 inner loops optimize regularization parameters.
To combine the results with the MSP baseline, as shown in \ref{tab:ridge}, we first train the Elastic-Net regression on the fold's train set, then perform a logistic regression on the same set using as features the scores from the Ridge model and the scores from the baseline. We then measure the AUROC on the fold's test set.

A pre-processing step to reduce the number of features is needed to reduce the computational cost of the Elastic-Net regression on $(L-1) \times D$ features. We keep the 100 most informative with respect to ANOVA F-values. We then use the following grid for selecting the best hyperparameters:
\begin{align*}
    l1\_ratio &= [0.9, 0.7, 0.5] \\
    C &= [0.01, 0.05, 0.1]
\end{align*}

\end{document}